\documentclass[journal]{IEEEtran}
\usepackage[bookmarks=true]{hyperref}
\usepackage{multicol}
\usepackage{graphicx}
\usepackage[colorinlistoftodos]{todonotes}
\usepackage{stfloats}
\usepackage{amsmath}
\usepackage{amssymb}
\usepackage{gensymb}
\usepackage{siunitx} 
\usepackage{mathtools}
\usepackage{commath}
\usepackage{multirow}
\usepackage{cite}

\usepackage{subcaption}

\ifCLASSINFOpdf
\else
\fi
\begin{document}
%
\title{Differential Spiral Joint Mechanism \\for Coupled Variable Stiffness Actuation}
%
%
%

\author{Mincheol~Kim,
        Ashish~D.~Deshpande,~\IEEEmembership{Member,~IEEE}
\thanks{All authors are with The University of Texas at Austin, TX 78705, USA.}
\thanks{Mincheol Kim and Ashish D. Deshpande are with the Department of Mechanical Engineering. e-mail: {\tt\small mincheol@utexas.edu, ashish@austin.utexas.edu}.}
\thanks{Manuscript received November 1, 2021.}}
\maketitle


\begin{abstract}
In this study, we present the Differential Spiral Joint (DSJ) mechanism for variable stiffness actuation in tendon-driven robots. The DSJ mechanism semi-decouples the modulation of position and mechanical stiffness, allowing independent trajectory tracking in different parameter space. Past studies show that increasing the mechanical stiffness achieves the wider range of renderable stiffness, whereas decreasing the mechanical stiffness improves the quality of actuator decoupling and shock absorbance. Therefore, it is often useful to modulate the mechanical stiffness to balance the required level of stiffness and safety. In addition, the DSJ mechanism offers a compact form factor, which is suitable for applications where the size and weight are important. The performance of the DSJ mechanism in various areas is validated through a set of experiments.
\end{abstract}


%
\IEEEpeerreviewmaketitle

\section{Introduction} \label{c1}
\begin{figure*}[b]
    \centering
    \subfloat[][\label{fig:nullfingers}]{\includegraphics{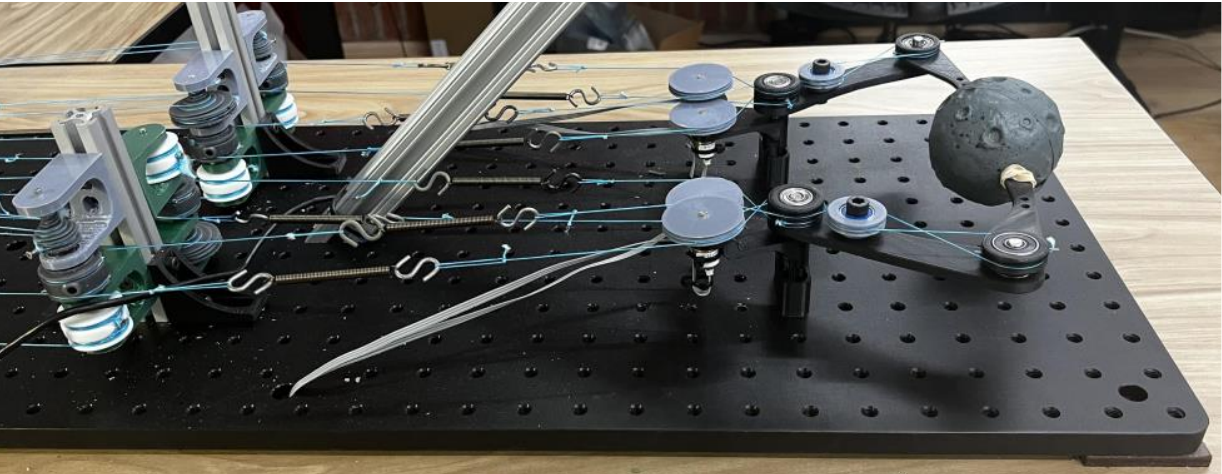}}%
    \subfloat[][\label{fig:realDSJ}]{\includegraphics{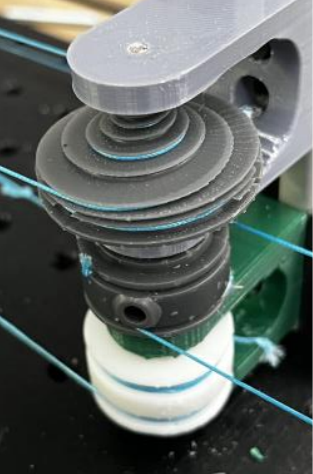}}%
    \caption{(a) Tendon-driven fingers are equipped with the Differential Spiral Joints. (b) A DSJ module provides passive stiffness modulation under compact settings.}
    \label{fig:overall_fig}
\end{figure*}

\IEEEPARstart{C}{urrent} robotic manipulation research focuses on not only the performance of the robot for certain tasks, but also on safety around the surrounding environment. This trend is well exhibited by the gradual shift from some of the early rigid robot hand designs \cite{tomovic1962adaptive, yamashita1963engineering, loucks1987modeling, jacobsen1986design} to the recently introduced compliant hands \cite{Zhou2018,alspach2018design, pedro2018closed, glick2018soft}. With proper hardware, studies show that it is possible to simulate compliant behavior through stiffness control with rigid robots as well \cite{Hogan1984, Salisbury1980, salisbury1982articulated, nguyen1989constructing}. However, certain behavior of the robot is heavily affected by the inherent stiffness of the hardware \cite{ning2020human, wolf2011dlr, garabini2011optimality}. Even under stiffness control, the apparent stiffness reverts back to the mechanical stiffness under high frequency excitations due to the limited control bandwidth \cite{Tagliamonte2014, vallery2008compliant}. For this reason, many studies emphasize the benefits of variable stiffness actuators with regards to safety and energy efficiency \cite{tonietti2005design, schiavi2008vsa}. 

Variable stiffness actuators employ complex mechanisms to mechanically change the stiffness, which provides certain benefits at the cost of structural complexity. This complexity may bring along undesirable characteristics such as large size, heavy weight, expensive cost, and complex control algorithms that hinder its application under practical settings. For example, there have been past attempts to employ variable stiffness actuators in robot hands \cite{Grebenstein2008, Butterfa2001}, but these existing approaches fail to avoid the aforementioned drawbacks. In the field of robotic manipulation under human-friendly settings, it is crucial to achieve the necessary level of safety while ensuring required performance. While lowering the actuator capacity is an effective way to reduce the overall weight and therefore improve safety, the output force or speed is reduced. Tendon-driven mechanisms allow separation of actuation from the driven link, which typically enables light and compact design of the robot links while maintaining the actuator performance. Therefore, employment of tendon-driven mechanism can be a safe and yet capable option for robotic fingers.

To this day, however, a compact and yet inexpensive variable stiffness actuator has yet to be introduced. In this paper, we propose the DSJ (Differential Spiral Joint) mechanism, which is a novel design for a compact variable stiffness actuator in tendon-driven mechanisms and experimentally validate its performance.

\section{Related Works} \label{c2}
Over the years, there have been many attempts in implementing variable stiffness actuation in robots. However, due to the size and weight constraints of robot hands, few of the existing approaches are feasible in practical applications. The simplest method is to use nonlinear springs in an antagonistic setup \cite{chiaradia2020nonlinear, nakanishi2011design, Kim2021}. Similar to our muscles, this approach relies on the physical characteristics of the nonlinear springs to progressively change its stiffness as it compresses and extends. However, the studies show that use of nonlinear springs for stiffness variation results in considerable energy loss due to the amount of force required for stiffness variation \cite{haddadin2011optimal, laurin1991design}. To overcome the energy loss while bringing the required performance of the fingers, the system may need actuators with the higher capacity, which increases the cost, size and weight of the overall system.

Another common approach is to vary the transmission ratio between the compensation force/torque from the compliant material and the deflection of the output link. This approach utilizes the relationship between the stiffness and the compensation force/torque caused by the deflection, and therefore generally requires complex mechanisms that are difficult to install in a compact environment such as in robot fingers. Most studies utilize the structure to nonlinearly guide the spring deflection, causing nonlinear compensatory force from linear springs \cite{hocaoglu2019design, wolf2011dlr, sun2018design, schiavi2008vsa}, which ultimately leads to variable stiffness. Due to the complex design in this approach, it is probable that the manufacturing is extremely difficult if it is even feasible, which significantly brings up the cost.

\begin{figure}[t]
    \centering
    \includegraphics[height=5.5cm]{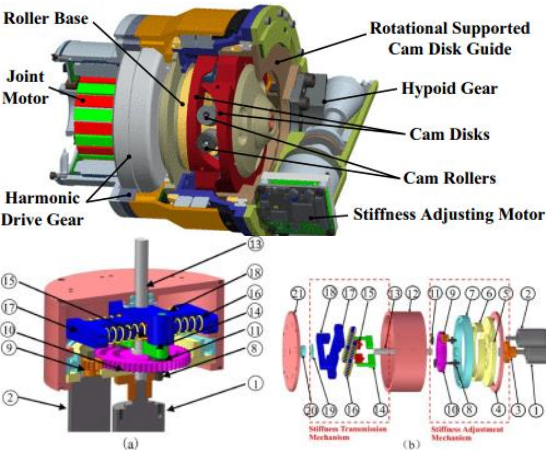}
    \caption{Some existing works are shown \cite{wolf2011dlr, sun2018design}. The working principle is that an actuator modulates the cam profiles, which are followed by the rollers that compress the springs depending on the amount of deviation from equilibrium.}
    \label{fig:other_works}
\end{figure}

Figure \ref{fig:other_works} shows a few examples of existing approaches. As shown in these works, it is extremely difficult to condense the actuator modules into compact space for the two main reasons: 1) the structural complexity that involves a variety of mechanical parts (e.g., bearings, linear rails, guiding structure, etc.) and 2) the size dependency of the entire module on the springs. The stiffness variation depends on the nonlinear transmission, which is dictated by travel of the springs. However, due to the size constraints, the travel length cannot be larger than a certain amount. To compensate for the small travel, the springs must be stronger, which leads to the larger springs, and in turn a larger design. This dilemma has caused this approach inapplicable in many of the compact applications.

The Differential Spiral Joint (DSJ) mechanism proposed in this study also employs the transmission variation method to modulate the physical stiffness of the actuation module. Unlike the existing approaches, however, the main advantage of the Differential Spiral Joint mechanism is its applicability in existing tendon-driven mechanisms without sacrificing the size, weight, or cost. Therefore, the use of the DSJ mechanism does not compromise the benefits of using tendon-driven mechanisms, while fully incorporating the important features of variable stiffness actuation. In the following chapters, we explore the design of the proposed DSJ mechanism, and validate its performance through a set of simulation and experiments.

\section{Differential Spiral Joint Mechanism} \label{c3}
Consider typical coupled, tendon-driven, series elastic fingers shown in Fig. \ref{fig:simple_finger}. The net joint torque from the tendons $\boldsymbol{\tau}$ is derived as following:

\begin{figure}[b]
    \centering\vspace{-3mm}
    \includegraphics{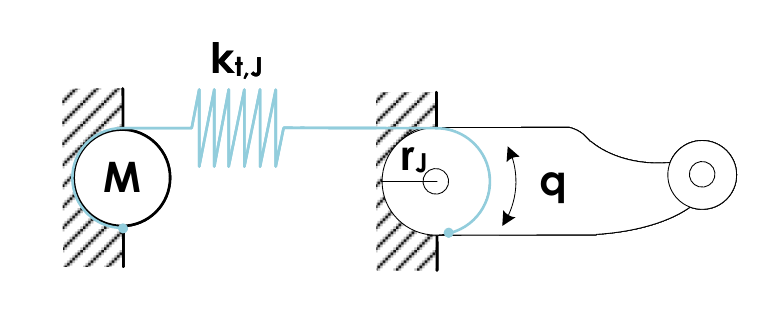}
    \caption{A simple schematic of a tendon-driven finger is shown. M refers to the actuator.}
    \label{fig:simple_finger}
\end{figure}

\begin{equation}
    \boldsymbol{\tau} = \mathbf{R}_J^T \mathbf{K}_{t,J} (\mathbf{L}_J - \mathbf{L}_m)
\end{equation}\label{eqn:1}
where $\mathbf{R}_J$, $\mathbf{K}_{t,J}$, $\mathbf{L}_J$, and $\mathbf{L}_m$ are the joint pulley radii, linear stiffness of the tendons, differences from equilibrium in tendon lengths caused by rotation of the joints or the motors, respectively. Then, the rotational stiffness of the joint $\mathbf{K}$ can be derived as following: 

\begin{equation}
    \begin{aligned}
        \mathbf{K} = \frac{\delta\boldsymbol{\tau}}{\delta \mathbf{q}} = \frac{\delta\mathbf{R}_J^T}{\delta \mathbf{q}} \mathbf{K}_{t,J}(\mathbf{L}_J - \mathbf{L}_m) + \mathbf{R}_J^T \mathbf{K}_{t,J}\frac{\delta\mathbf{L}_J}{\delta \mathbf{q}}
    \end{aligned}\label{eqn:2}
\end{equation}

Since $\mathbf{L}_J$ is directly related to $\mathbf{R}_J$, there are two ways to change the stiffness of the mechanism: 1) by changing the spring stiffness directly ($\mathbf{K}_{t,J}$) or 2) by changing the transmission ratio ($\mathbf{R}_J$), which is related with the joint radius. However, as mentioned in Sec. \ref{c2}, most approaches that aim to change the spring stiffness are inefficient in terms of energy consumption. Furthermore, in typical tendon-driven systems, $\mathbf{R}_J$ stays fixed, due to the inability to change the radius of the joint that is rigidly connected to the robot link \cite{Kim2021}. In this work, we propose the Differential Spiral Joint mechanism to modulate the passive stiffness and drastically improve the energy efficiency.

\subsection{Structure}\label{c3s1}
\begin{figure}[t]
    \centering
    \includegraphics{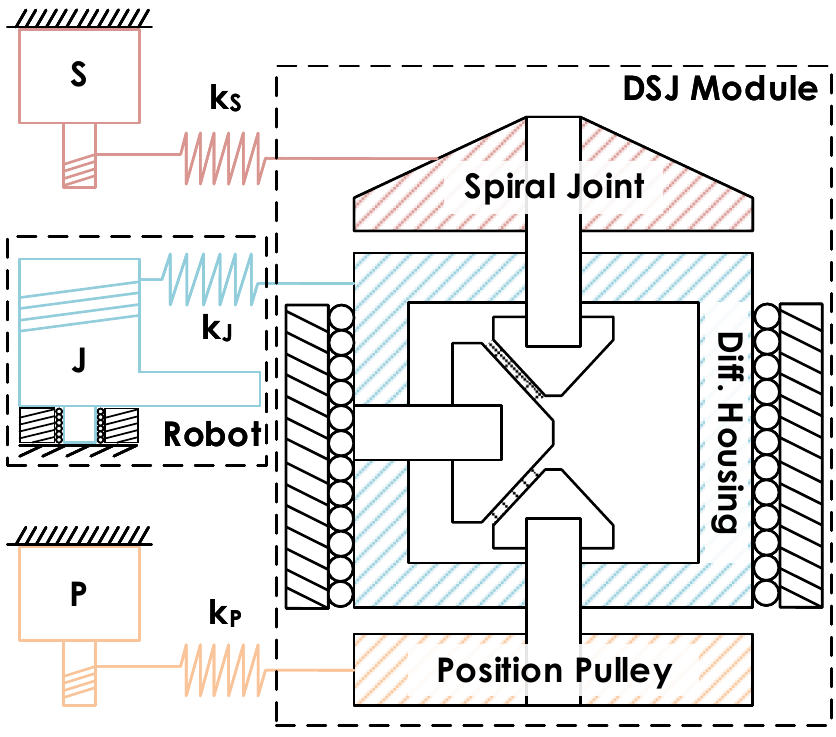}
    \caption{The DSJ mechanism consists of three primary connections. The spiral joint (red) is connected to the stiffness-related actuator (S), the differential housing (blue) is connected to the robot joint (J), and the position pulley (orange) is connected to the position-related actuator (P). The spiral joint, the differential housing, and the position pulley are connected via pinion gears in a differential gearbox. This assembly is referred to as a DSJ module.}
    \label{fig:DSJ_schematic}
\end{figure}

The overall structure of the DSJ mechanism is shown in Fig. \ref{fig:DSJ_schematic}. The configuration of the spiral joint is modulated by the stiffness-related actuator, whereas the joint position is modulated using the position-related actuator. These two actuators are connected to the robot through the differential gear assembly, hence the name Differential Spiral Joint mechanism. This structure allows the position and the stiffness of the joint to be semi-decoupled, where only the position-related actuator needs to move to adjust the position of the joint but both actuators need to move accordingly to adjust the stiffness. 

\begin{figure*}[t]
    \centering
    \includegraphics{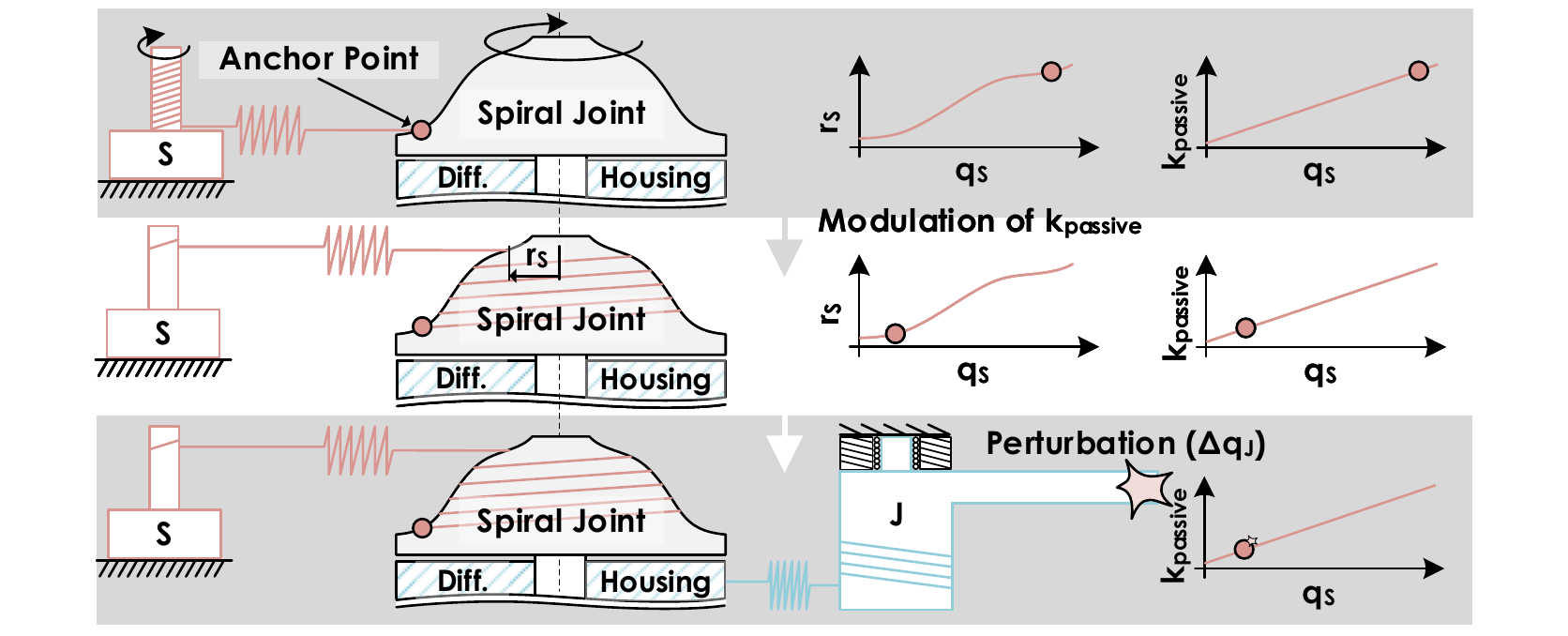}
    \caption{As the spiral joint is rotated, the overall passive stiffness is modulated.}
    \label{fig:DSJ_rotation}
\end{figure*}

The spiral joint is synthesized according to the desired moment arm $r_s$, which varies with the configuration of the spiral joint, $q_s$ (i.e., $r_s=r_s(q_s)$). Owing to the mechanical structure, torques caused by the deviation of the link from equilibrium are evenly distributed to the position- and stiffness-related tendons, and to the joint-related tendons after amplification. As a result, the DSJ module acts a set of torsional springs connected in series. This unique structure allows energy-efficient modulation of the passive stiffness as shown in Fig. \ref{fig:DSJ_rotation}. The overall rotational stiffness of the robot $\mathbf{K}_\text{passive}$ can be calculated as:

\begin{equation}
    \begin{aligned}
        \mathbf{K}_\text{passive}^{-1} &= \mathbf{K}_{P,j}^{-1} + \mathbf{K}_{S,j}^{-1} + \mathbf{K}_{J,j}^{-1} \\
        \mathbf{K}_\text{passive}^{-1} &= n^{-2}\mathbf{K}_{P,d}^{-1} + n^{-2}\mathbf{K}_{S,d}^{-1} + \mathbf{K}_{J,j}^{-1}
    \end{aligned}\label{eqn:3}
\end{equation}
where $\mathbf{K}_{P,j}$, $\mathbf{K}_{S,j}$, and $\mathbf{K}_{J,j}$ are the joint-level stiffness caused by the springs in the position-, stiffness-, and joint-related tendons, which have linear stiffness of $k_p$, $k_s$, and $k_j$ respectively. $\mathbf{K}_{P,d}$ and $\mathbf{K}_{S,d}$ refer to differential-housing-level stiffness, and $n$ is the ratio from the radius of the differential housing to that of the joints (amplification ratio).

Since the spiral joint has a varying radius about its rotation $q_s$, the joint stiffness of the robot $\mathbf{K}_\text{passive}$ can be determined as a function of $q_s$ as well. In other words, we can \emph{shape} the spiral joint to match the desired stiffness profile $\mathbf{K}_\text{passive}(q_s)$ of the robot. Rearranging Eq.\eqref{eqn:3} about $\mathbf{K}_{S,j}$ gives us:

\begin{equation}
    \begin{aligned}
        \mathbf{K}_{S,j} = (\mathbf{K}_\text{passive}^{-1} - n^{-2}\mathbf{K}_{P,d}^{-1} - \mathbf{K}_{J,j}^{-1})^{-1}
    \end{aligned}\label{eqn:4}
\end{equation}
where $0<\mathbf{K}_\text{passive}<(n^{-2}\mathbf{K}_{P,d}^{-1} +\mathbf{K}_{J,j}^{-1})^{-1}$ must hold. To conveniently calculate the desired radius of the spiral joints $\mathbf{R}_S$, we make the following assumptions about the spiral joints:

\begin{itemize}
    \item Change in the radius $\mathbf{R}_S$ about the rotation $\mathbf{q}_S$ is negligible compared to the radius (i.e., $\frac{\delta\mathbf{R}_S}{\delta\mathbf{q}_S} << \mathbf{R}_S$).
    \item Change in the vertical elevation of the grooves $\mathbf{Z}_S$ about the rotation $\mathbf{q}_S$ is negligible compared to the radius (i.e., $\frac{\delta\mathbf{Z}_S}{\delta\mathbf{q}_S} << \mathbf{R}_S$).
\end{itemize}
These assumptions are achievable by adjusting the operating range of the spiral joint $\mathbf{q}_S$ and the vertical elevation $\mathbf{Z}_S$, which are independent from $\mathbf{R}_S$. Then, the length of the tendon wrapping around the spiral joint in 3D space, $\mathbf{L}_S$ can be simplified (i.e., $\frac{\delta\mathbf{L}_S}{\mathbf{q}_S} = \sqrt{(\frac{\delta\mathbf{R}_S}{\mathbf{q}_S})^2 + \mathbf{R}_S^2 + (\frac{\delta\mathbf{Z}_S}{\mathbf{q}_S})^2} \approx \mathbf{R}_S$). Combining this knowledge with Eq. \eqref{eqn:2}, the instantaneous joint-level stiffness can be found:

\begin{equation}
    \begin{aligned}
        \mathbf{K}_{S,j} = (\mathbf{R}_S\mathbf{R}_D^{-1}\mathbf{R}_J)^T \mathbf{K}_{t,S}\mathbf{R}_S\mathbf{R}_D^{-1}\mathbf{R}_J
    \end{aligned}\label{eqn:5}
\end{equation}
where $\mathbf{R}_D$ and $\mathbf{R}_J$ are the pulley radii for the differential housing and the joints, respectively. The first term in Eq. \eqref{eqn:2} disappears because the joint is at equilibrium prior to the perturbation (i.e., $\mathbf{L}_J = \mathbf{L}_m$). Therefore, $\mathbf{R}_S$ can be acquired by combining Eq. \eqref{eqn:4} and \eqref{eqn:5}.

\begin{figure}[t]
    \centering
    \includegraphics[width=8.5cm]{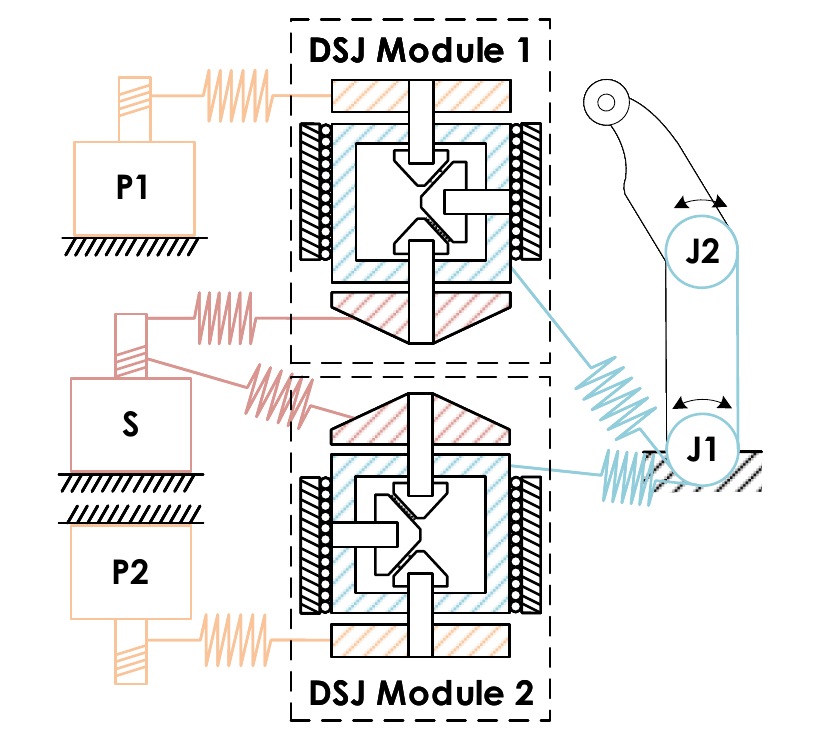}
    \caption{Multiple DSJ modules can be used simultaneously for multi-DoF tendon-driven robots. Note that one stiffness-related actuator dictates the spiral joints in both modules, whereas the robot joints are controlled individually using two position-related actuators.}
    \label{fig:DSJ_extension}
\end{figure}

\subsection{Multi-DoF and Coupled Stiffness Example}\label{c3s3}
Assume a coupled, tendon-driven 2-DoF finger as shown in Fig. \ref{fig:DSJ_extension}, which has the following radius matrices from its structure:

\begin{figure}[t]
    \centering
    \includegraphics[width=8.5cm]{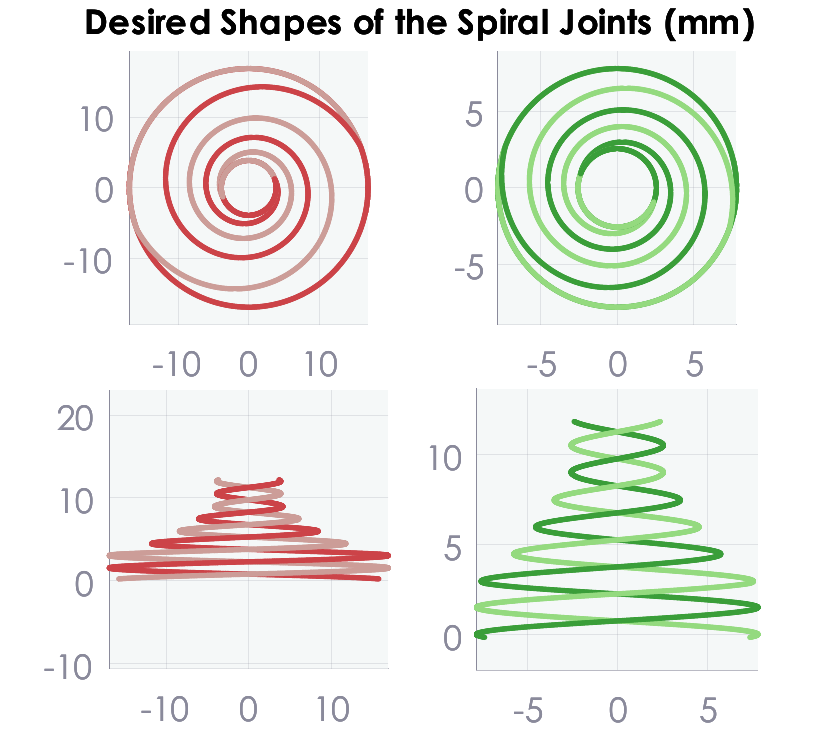}
    \caption{The resulting spiral profiles for both joints from top and side views.}
    \label{fig:spiraljoint}
\end{figure}

\begin{equation}
    \begin{aligned}
        \mathbf{R}_J = 
        \left[\begin{matrix}
        r_j & 0\\
        r_j & r_j
        \end{matrix}\right],\hfill
        \mathbf{R}_D = 
        \left[\begin{matrix}
        r_d & 0\\
        0 & r_d
        \end{matrix}\right],\hfill
        \mathbf{R}_S = 
        \left[\begin{matrix}
        r_{s1} & 0\\
        0 & r_{s2}
        \end{matrix}\right]
    \end{aligned}\label{eqn:6}
\end{equation}

Combining Eq. \eqref{eqn:5} and \eqref{eqn:6}, we get:
\begin{equation}
    \begin{aligned}
        \mathbf{K}_{S,j} = 
        \left[\begin{matrix}
        n^2k_s(r_{s1}^2+r_{s2}^2) & n^2k_sr_{s2}^2\\
        n^2k_sr_{s2}^2 & n^2k_sr_{s2}^2
        \end{matrix}\right]
    \end{aligned}\label{eqn:7}
\end{equation}
Therefore, $r_{s1}$ and $r_{s2}$ can be calculated from Eq. \eqref{eqn:4}-\eqref{eqn:7}. For example, if $k_s=k_j=k_p = 875.63 N/m$, $r_j=r_d = 12 mm$, $n=1$, linearly varying $\mathbf{K}_\text{passive} \in [0.2\mathbf{K}_\text{max}, 0.8\mathbf{K}_\text{max}]$, and $z_s \in [0, 12]mm$ about $q_s\in [0,4\pi]$ are desired, $r_{s1}$ and $r_{s2}$ can be uniquely determined. Note that $\mathbf{K}_\text{max} \triangleq (n^{-2}\mathbf{K}_{P,d}^{-1} +\mathbf{K}_{J,j}^{-1})^{-1}$ is the upper limit of the achievable passive stiffness for this mechanical setup, which is when $\mathbf{K}_{S,j} \rightarrow \infty$.

\begin{figure}[t]
    \centering
    \includegraphics[width=8.5cm]{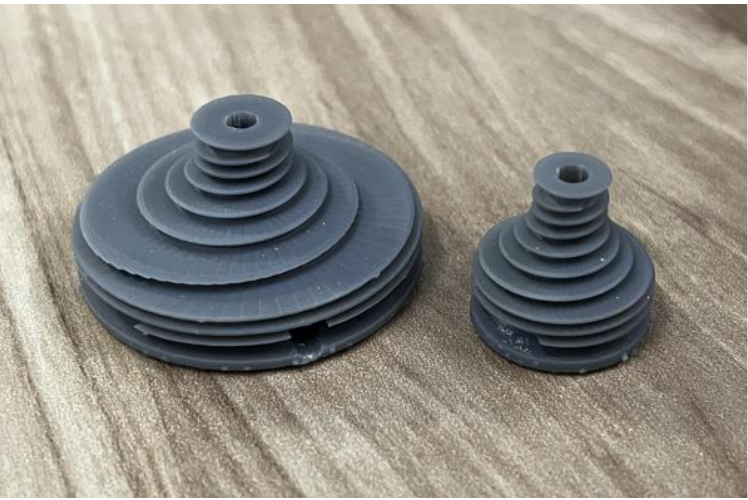}
    \caption{The manufactured spiral joints are shown. Tough 2000 resin is used for the spiral joints in this study.}
    \label{fig:spiraljoint_real}
\end{figure}

\begin{figure}[b]
    \centering
    \includegraphics[height=4.5cm]{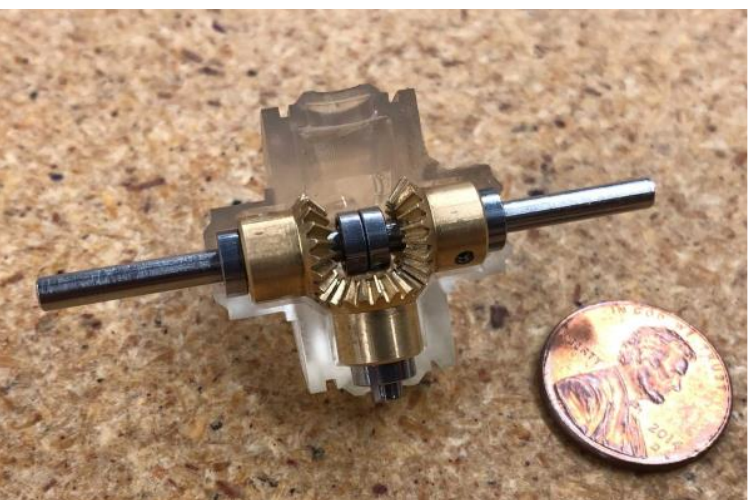}
    \caption{The compact differential module connects the three essential components- position and stiffness actuators and the robot link. The differential module comes in a small form factor that is comparable to a penny.}
    \label{fig:mini_dsj}
\end{figure}

Considering the radius of the tendons to be used in the experiment, an appropriate 3D model of the spiral joint can be chosen. For this study, we make grooves to work with $0.5mm$ braided tendons. The resulting spiral joints are shown in Fig. \ref{fig:spiraljoint}. Notice that the spiral grooves are symmetrically made to enable the antagonistic setup. This particular design passively ensures that the moment arm acting at any given time is identical for both tendons. It is important to note that these particular spiral joints provide the desired scaled range of the maximum passive stiffness $\mathbf{K}_\text{max}$ for any linear spring stiffness $k_s, k_j, k_p$ as long as they maintain the same ratio between them, leading to versatile applicability. The resulting spiral joints are 3D printed using Tough 2000 resin, and are shown in Fig. \ref{fig:spiraljoint_real}. These spiral joints are placed on one of the shafts of the differential module shown in Fig. \ref{fig:mini_dsj}.

Owing to the configuration-dependency of spiral joints, one actuator is capable of controlling the entire passive stiffness of the fingers using identical spiral joints fixed to its shaft. This strategy allows 1:1 mapping from the actuator rotation to all of the spiral joint rotations, $q_s$, which allows simple feedforward control. The resulting setup is shown in Fig. \ref{fig:realDSJ}.

\begin{figure}[t]
    \centering
    \includegraphics{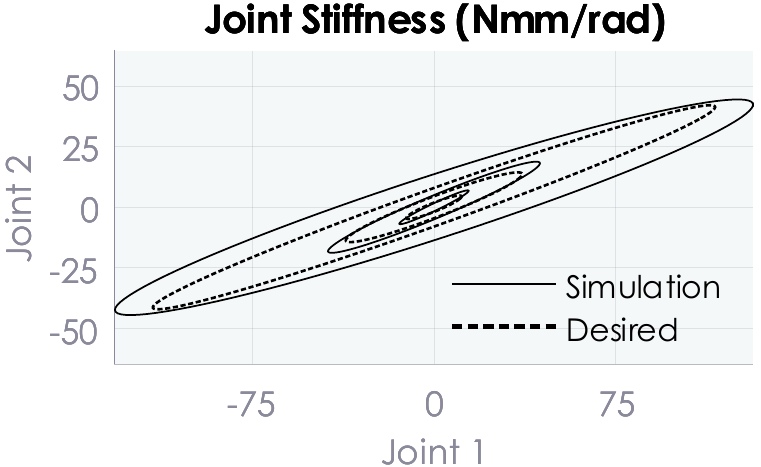}
    \caption{Three different joint stiffness ellipses are shown.}
    \label{fig:coupled_stiffness}
\end{figure}

The desired and simulated joint stiffness ellipses at three different configurations of the spiral joints are plotted in Fig. \ref{fig:coupled_stiffness}. The stiffness is linearly regressed from the compensatory torques from uniform joint-level deviation (i.e., a 2D circle with a radius of 20\degree). A slight mismatch from the desired stiffness ellipses is caused by the assumptions in Sec. \ref{c3s1} and varying instantaneous stiffness about the rotation of the spiral joints. This error can also be contributed by the nonzero deviation, which was neglected in the formulation of the spiral joints for simplicity. Based on these results, we postulate that the DSJ mechanism successfully accommodates the desired mechanical stiffness with some reasonable error.

\section{Validation} \label{c4}

\begin{figure}[b]
    \centering
    \includegraphics{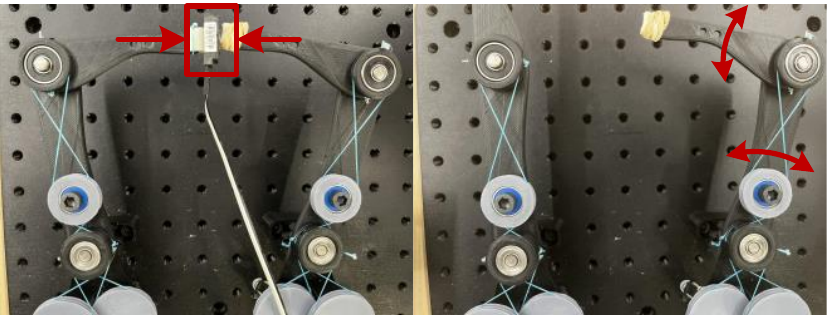}
    \caption{(a) Fingers grasp a force sensor (inside the red box) and close the fingers gradually to increase the deviation in the grasp direction with varying stiffness. (b) Joint positions are independently modulated within a single finger.}
    \label{fig:experiment_setup}
\end{figure}

In this section, we verify the performance of the Differential Spiral Joint mechanism in its essential aspects: 1) stiffness control and 2) position control. For all experiments, a 2-DoF system is used, where joints are coupled via tendons. Note that the stiffness for each joint can be individually modulated, but for mechanical simplicity, we modulate stiffness for all joints simultaneously. Furthermore, we examine its performance in task space as it is difficult to implement joint torque sensors into small finger joints. Therefore, we let the fingers grasp the force sensor as shown in Fig. \ref{fig:experiment_setup}, and analyze the force readings in comparison to the desired force measurements.

\subsection{Stiffness modulation} \label{c4s1}
As previously described, the DSJ mechanism varies the joint stiffness by adjusting the transmission ratio between the elastic deformation and the output link deformation. A common way of assessing the joint stiffness is to measure the resulting force/torque while varying the deflection of the output link. Due to the difficulty of installing joint torque sensors in the compact setup, we utilize load cell-based sensors to measure forces acting on the end-effector, and calculate the desired compensatory forces based on the task space stiffness model. The task space stiffness involves derivatives of different Jacobians \cite{Kim2021}. Consider the definition of joint stiffness $\mathbf{K}$:

\begin{equation}\begin{split}\label{eqn:8}
    \mathbf{K} &= \frac{\delta\boldsymbol{\tau}}{\delta \mathbf{q}} = \frac{\delta}{\delta \mathbf{q}}(\mathbf{J}^T\mathbf{J}^T_p\mathbf{F}_p)\\
    &= \frac{\delta\mathbf{J}^T}{\delta \mathbf{q}}\mathbf{J}^T_p\mathbf{F}_p + \mathbf{J}^T\frac{\delta \mathbf{J}^T_p}{\delta \mathbf{q}}\mathbf{F}_p + \mathbf{J}^T\mathbf{J}^T_p\frac{\delta \mathbf{F}_p}{\delta \mathbf{q}}\\
    &= \frac{\delta\mathbf{J}^T}{\delta \mathbf{q}}\mathbf{J}^T_p\mathbf{F}_p + \mathbf{J}^T\frac{\delta \mathbf{J}^T_p}{\delta \mathbf{x}}\mathbf{J}\mathbf{F}_p + \mathbf{J}^T\mathbf{J}^T_p\mathbf{K}_p\mathbf{J}_p\mathbf{J}
\end{split}\end{equation}
where the subscript $p$, $\mathbf{x}$, and $\mathbf{q}$ denote the task, Cartesian, and joint space variables, respectively. $\mathbf{J}$ is a Jacobian and $\mathbf{F}_p$ refers to the task space force. Therefore, the expected task space stiffness from the model, $\mathbf{K}_p$, can be found as:
\begin{equation}\label{eqn:9}
    \mathbf{K}_p = \mathbf{J}^{+T}_p\mathbf{J}^{+T}(\mathbf{K} - (\frac{\delta\mathbf{J}^T}{\delta \mathbf{q}}\mathbf{J}^T_p + \mathbf{J}^T\frac{\delta \mathbf{J}^T_p}{\delta \mathbf{x}}\mathbf{J})\mathbf{F}_p)\mathbf{J}^+\mathbf{J}^+_p
\end{equation}

\begin{figure*}[t]
    \centering
    \includegraphics{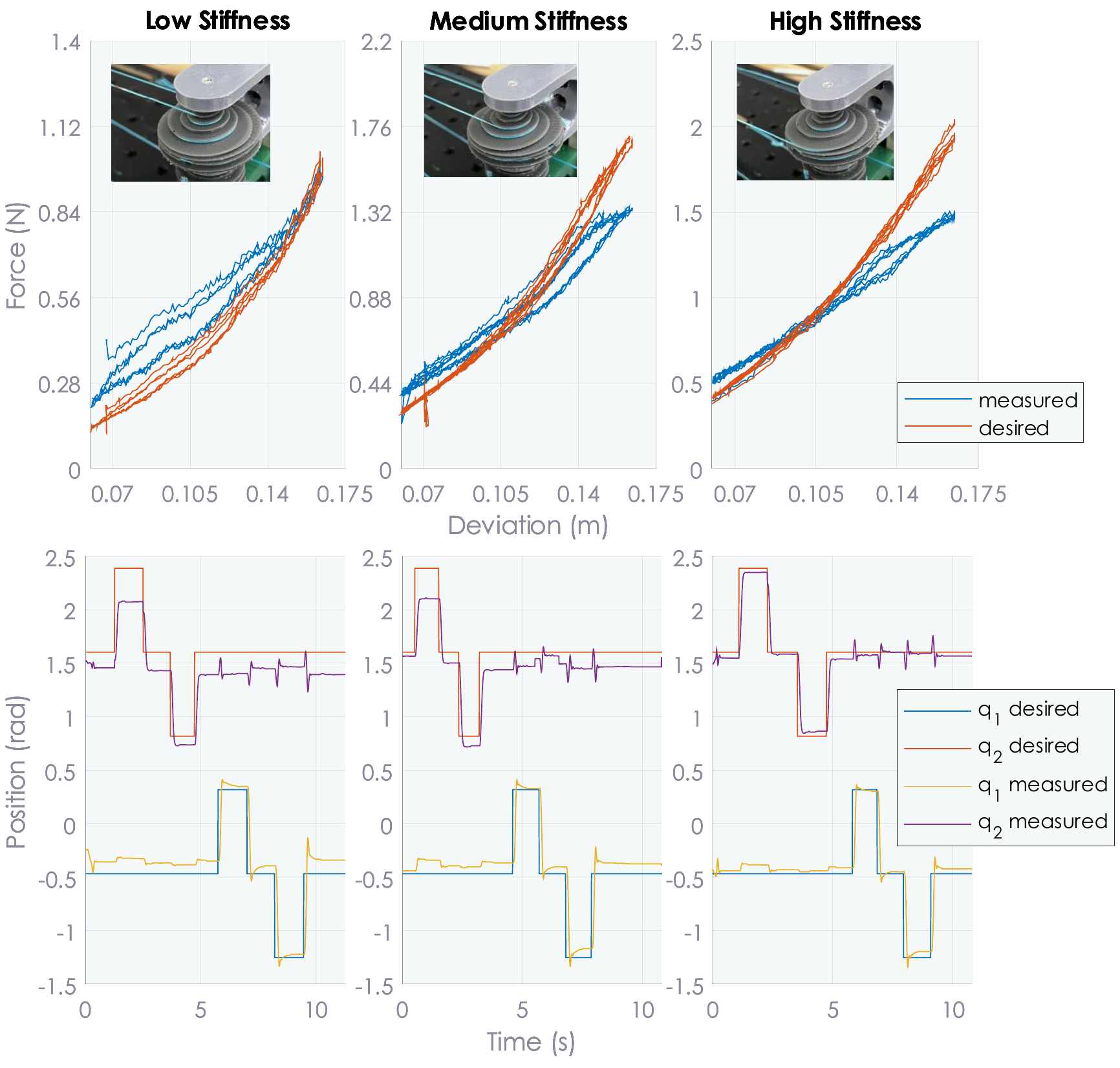}
    \caption{(Top) Desired and measured compensatory forces with varying deviation from equilibrium are shown. As can be seen from these plots, it can be confirmed that the stiffness is tracked with reasonable fidelity under practical deflection despite the continuous transmission modulation. A slight mismatch is inevitable as the nonlinear friction is not accounted for and the stiffness of the springs is assumed to be linear across the operating range. (Bottom) Joint positions can be independently modulated with constant joint stiffness. Different properties can be observed depending on the mechanical stiffness.}
    \label{fig:stiffnessmodulation}
\end{figure*}
where $+$ refers to the generalized inverse of a matrix. Notice that there is a force-dependent term in Eq. \eqref{eqn:9}, which is defined in task space (i.e., $\mathbf{F}_p = \mathbf{K}_p\Delta \mathbf{p}$). $\Delta \mathbf{p}$ contains the deviation in the grasping direction. This results in the increased stiffness as the grasping force increases, as shown in Fig. \ref{fig:stiffnessmodulation}. From these results, we conclude that the proposed DSJ mechanism modulates the passive stiffness as intended with some reasonable errors at extreme deviations ($>14cm$) for higher stiffness configuration. We postulate that these errors occur from the discrepancy between the model and the manufactured robot. 

\subsection{Position modulation} \label{c4s2}
We also examine the performance of the DSJ mechanism in position modulation. Specifically, we examine the step response of the robot under constant but different joint stiffness values. In this experiment, we vary the first and second joint positions of one finger independently, denoted as $q_1$ and $q_2$. The measured joint positions based on the encoder readings are then compared to the commanded joint positions.

The results are shown in the bottom of Fig. \ref{fig:stiffnessmodulation}. Here, we can observe important physical properties of a typical VSA. The response profile depends on the stiffness, which affects the amount of oscillations and the settling time. Lower stiffness results in lower frequency oscillations and slower settling time, and vice-versa for the higher stiffness. The lower stiffness is more conforming to the external disturbances, hence improved safety. The higher stiffness results in more accurate trajectory tracking, which is required in some tasks. Position modulation under varying stiffness is frequently used in robotics \cite{Salisbury1980, Hogan1984}. We conclude from these results that both position and stiffness can be independently modulated through the DSJ mechanism.

\section{Conclusion} \label{c5}
In this study, we introduce the Differential Spiral Joint (DSJ) mechanism, which is a novel variable stiffness actuator that utilizes a unique spiral joint design that semi-decouples the stiffness modulation from position modulation using a differential mechanism. The proposed mechanism is designed to be used in remote systems that require a compact form factor, which is main benefit of tendon-driven mechanisms. Therefore, we anticipate the DSJ mechanism to be useful in dexterous manipulation by accommodating varying dynamic properties. Despite the compact design, the DSJ mechanism successfully provides independent control over stiffness and position, and many other benefits that are offered by typical VSAs. In future work, we plan to extend the study to dexterous manipulation, where the variable physical stiffness can significantly improve the performance of the fingers.



\section*{Acknowledgment}
This work is supported by xxx. The authors would also like to thank Ashwin Hingwe for his help in manufacturing the parts used in this study.

\ifCLASSOPTIONcaptionsoff
  \newpage
\fi



%
\bibliographystyle{IEEEtran}
\bibliography{IEEEfull,DSJ}

\begin{thebibliography}{10}
\providecommand{\url}[1]{#1}
\csname url@samestyle\endcsname
\providecommand{\newblock}{\relax}
\providecommand{\bibinfo}[2]{#2}
\providecommand{\BIBentrySTDinterwordspacing}{\spaceskip=0pt\relax}
\providecommand{\BIBentryALTinterwordstretchfactor}{4}
\providecommand{\BIBentryALTinterwordspacing}{\spaceskip=\fontdimen2\font plus
\BIBentryALTinterwordstretchfactor\fontdimen3\font minus
  \fontdimen4\font\relax}
\providecommand{\BIBforeignlanguage}[2]{{%
\expandafter\ifx\csname l@#1\endcsname\relax
\typeout{** WARNING: IEEEtran.bst: No hyphenation pattern has been}%
\typeout{** loaded for the language `#1'. Using the pattern for}%
\typeout{** the default language instead.}%
\else
\language=\csname l@#1\endcsname
\fi
#2}}
\providecommand{\BIBdecl}{\relax}
\BIBdecl

\bibitem{tomovic1962adaptive}
R.~Tomovic and G.~Boni, ``An adaptive artificial hand,'' \emph{IRE Transactions
  on Automatic Control}, vol.~7, no.~3, pp. 3--10, 1962.

\bibitem{yamashita1963engineering}
T.~Yamashita, ``Engineering approaches to function of fingers,'' \emph{Report
  of the Institute of Industrial Science the Univ. of Tokyo}, vol.~13, no.~3,
  pp. 60--110, 1963.

\bibitem{loucks1987modeling}
C.~Loucks, V.~Johnson, P.~Boissiere, G.~Starr, and J.~Steele, ``Modeling and
  control of the stanford/jpl hand,'' in \emph{Proceedings. 1987 IEEE
  International Conference on Robotics and Automation}, vol.~4.\hskip 1em plus
  0.5em minus 0.4em\relax IEEE, 1987, pp. 573--578.

\bibitem{jacobsen1986design}
S.~Jacobsen, E.~Iversen, D.~Knutti, R.~Johnson, and K.~Biggers, ``Design of the
  utah/mit dextrous hand,'' in \emph{Proceedings. 1986 IEEE International
  Conference on Robotics and Automation}, vol.~3.\hskip 1em plus 0.5em minus
  0.4em\relax IEEE, 1986, pp. 1520--1532.

\bibitem{Zhou2018}
J.~Zhou, J.~Yi, X.~Chen, Z.~Liu, and Z.~Wang,
  ``\href{https://ieeexplore.ieee.org/abstract/document/8399511}{BCL-13: A
  13-DOF Soft robotic hand for dexterous grasping and in-hand manipulation.}''
  \emph{IEEE Robotics and Automation Letters}, vol.~3, pp. 3379--3386, 2018.

\bibitem{alspach2018design}
A.~Alspach, J.~Kim, and K.~Yamane, ``Design and fabrication of a soft robotic
  hand and arm system,'' in \emph{2018 IEEE International Conference on Soft
  Robotics (RoboSoft)}.\hskip 1em plus 0.5em minus 0.4em\relax IEEE, 2018, pp.
  369--375.

\bibitem{pedro2018closed}
P.~Pedro, C.~Ananda, P.~Rafael, A.~Carlos, and B.~Alexandre, ``Closed structure
  soft robotic gripper,'' in \emph{2018 IEEE International Conference on Soft
  Robotics (RoboSoft)}.\hskip 1em plus 0.5em minus 0.4em\relax IEEE, 2018, pp.
  66--70.

\bibitem{glick2018soft}
P.~Glick, S.~A. Suresh, D.~Ruffatto, M.~Cutkosky, M.~T. Tolley, and A.~Parness,
  ``A soft robotic gripper with gecko-inspired adhesive,'' \emph{IEEE Robotics
  and Automation Letters}, vol.~3, no.~2, pp. 903--910, 2018.

\bibitem{Hogan1984}
N.~Hogan, ``Adaptive control of mechanical impedance by coactivation of
  antagonist muscles,'' \emph{IEEE Transactions on automatic control}, pp.
  681--690, 1984.

\bibitem{Salisbury1980}
J.~Salisbury,
  ``\href{https://ieeexplore.ieee.org/abstract/document/4046624}{Active
  stiffness control of a manipulator in cartesian coordinates.}'' \emph{19th
  IEEE conference on decision and control including the symposium on adaptive
  processes}, 1980.

\bibitem{salisbury1982articulated}
J.~K. Salisbury and J.~J. Craig, ``Articulated hands: Force control and
  kinematic issues,'' \emph{The International journal of Robotics research},
  vol.~1, no.~1, pp. 4--17, 1982.

\bibitem{nguyen1989constructing}
V.-D. Nguyen, ``Constructing stable grasps,'' \emph{The International Journal
  of Robotics Research}, vol.~8, no.~1, pp. 26--37, 1989.

\bibitem{ning2020human}
Y.~Ning, Y.~Liu, F.~Xi, K.~Huang, and B.~Li, ``Human-robot interaction control
  for robot driven by variable stiffness actuator with force self-sensing,''
  \emph{IEEE Access}, 2020.

\bibitem{wolf2011dlr}
S.~Wolf, O.~Eiberger, and G.~Hirzinger, ``The dlr fsj: Energy based design of a
  variable stiffness joint,'' in \emph{2011 IEEE International Conference on
  Robotics and Automation}.\hskip 1em plus 0.5em minus 0.4em\relax IEEE, 2011,
  pp. 5082--5089.

\bibitem{garabini2011optimality}
M.~Garabini, A.~Passaglia, F.~Belo, P.~Salaris, and A.~Bicchi, ``Optimality
  principles in variable stiffness control: The vsa hammer,'' in \emph{2011
  Ieee/Rsj International Conference on Intelligent Robots and Systems}.\hskip
  1em plus 0.5em minus 0.4em\relax IEEE, 2011, pp. 3770--3775.

\bibitem{Tagliamonte2014}
N.~Tagliamonte and D.~Accoto,
  ``\href{https://journals.sagepub.com/doi/abs/10.1177/0959651813511615}{Passivity
  constraints for the impedance control of series elastic actuators},''
  \emph{Proceedings of the Institution of Mechanical Engineers, Part I: Journal
  of Systems and Control Engineering}, vol. 228, no.~3, pp. 138--153, 2014.

\bibitem{vallery2008compliant}
H.~Vallery, J.~Veneman, E.~Van~Asseldonk, R.~Ekkelenkamp, M.~Buss, and H.~Van
  Der~Kooij, ``Compliant actuation of rehabilitation robots,'' \emph{IEEE
  Robotics \& Automation Magazine}, vol.~15, no.~3, pp. 60--69, 2008.

\bibitem{tonietti2005design}
G.~Tonietti, R.~Schiavi, and A.~Bicchi, ``Design and control of a variable
  stiffness actuator for safe and fast physical human/robot interaction,'' in
  \emph{Proceedings of the 2005 IEEE international conference on robotics and
  automation}.\hskip 1em plus 0.5em minus 0.4em\relax IEEE, 2005, pp. 526--531.

\bibitem{schiavi2008vsa}
R.~Schiavi, G.~Grioli, S.~Sen, and A.~Bicchi, ``Vsa-ii: A novel prototype of
  variable stiffness actuator for safe and performing robots interacting with
  humans,'' in \emph{2008 IEEE International Conference on Robotics and
  Automation}.\hskip 1em plus 0.5em minus 0.4em\relax IEEE, 2008, pp.
  2171--2176.

\bibitem{Grebenstein2008}
M.~Grebenstein and P.~van~der Smagt,
  ``\href{https://www.tandfonline.com/doi/abs/10.1163/156855308X291836}{Antagonism
  for a highly anthropomorphic hand–arm system},'' \emph{Advanced Robotics},
  vol.~22, pp. 39--55, 2008.

\bibitem{Butterfa2001}
J.~Butterfaß, M.~Grebenstein, H.~Liu, and G.~Hirzinger,
  ``\href{https://ieeexplore.ieee.org/abstract/document/932538}{DLR-Hand II:
  Next generation of a dextrous robot hand.}'' \emph{IEEE International
  Conference on Robotics and Automation}, pp. 109--114, 2001.

\bibitem{chiaradia2020nonlinear}
D.~Chiaradia, L.~Tiseni, D.~Leonardis, and A.~Frisoli, ``Nonlinear
  characterization of a compact series visco-elastic element for tendon-driven
  actuation,'' in \emph{The International Conference of IFToMM ITALY}.\hskip
  1em plus 0.5em minus 0.4em\relax Springer, 2020, pp. 378--385.

\bibitem{nakanishi2011design}
Y.~Nakanishi, N.~Ito, T.~Shirai, M.~Osada, T.~Izawa, S.~Ohta, J.~Urata,
  K.~Okada, and M.~Inaba, ``Design of powerful and flexible musculoskeletal arm
  by using nonlinear spring unit and electromagnetic clutch opening
  mechanism,'' in \emph{2011 11th IEEE-RAS International Conference on Humanoid
  Robots}.\hskip 1em plus 0.5em minus 0.4em\relax IEEE, 2011, pp. 377--382.

\bibitem{Kim2021}
M.~Kim and A.~Deshpande, ``Balancing stability and stiffness through the
  optimization of parallel compliance: Using coupled tendon routing,''
  \emph{IEEE Robotics Automation Magazine}, vol.~28, no.~2, pp. 14--22, 2021.

\bibitem{haddadin2011optimal}
S.~Haddadin, M.~Weis, S.~Wolf, and A.~Albu-Sch{\"a}ffer, ``Optimal control for
  maximizing link velocity of robotic variable stiffness joints,'' \emph{IFAC
  Proceedings Volumes}, vol.~44, no.~1, pp. 6863--6871, 2011.

\bibitem{laurin1991design}
K.~F. Laurin-Kovitz, J.~E. Colgate, and S.~D. Carnes, ``Design of components
  for programmable passive impedance.'' in \emph{ICRA}, vol.~2, 1991, pp.
  1476--1481.

\bibitem{hocaoglu2019design}
E.~Hocaoglu and V.~Patoglu, ``Design, implementation and evaluation of a
  variable stiffness transradial hand prosthesis,'' \emph{arXiv preprint
  arXiv:1910.12569}, 2019.

\bibitem{sun2018design}
J.~Sun, Z.~Guo, D.~Sun, S.~He, and X.~Xiao, ``Design, modeling and control of a
  novel compact, energy-efficient, and rotational serial variable stiffness
  actuator (svsa-ii),'' \emph{Mechanism and Machine Theory}, vol. 130, pp.
  123--136, 2018.

\end{thebibliography}

%

\begin{IEEEbiography}[{\includegraphics[width=1in,height=1.25in,clip,keepaspectratio]{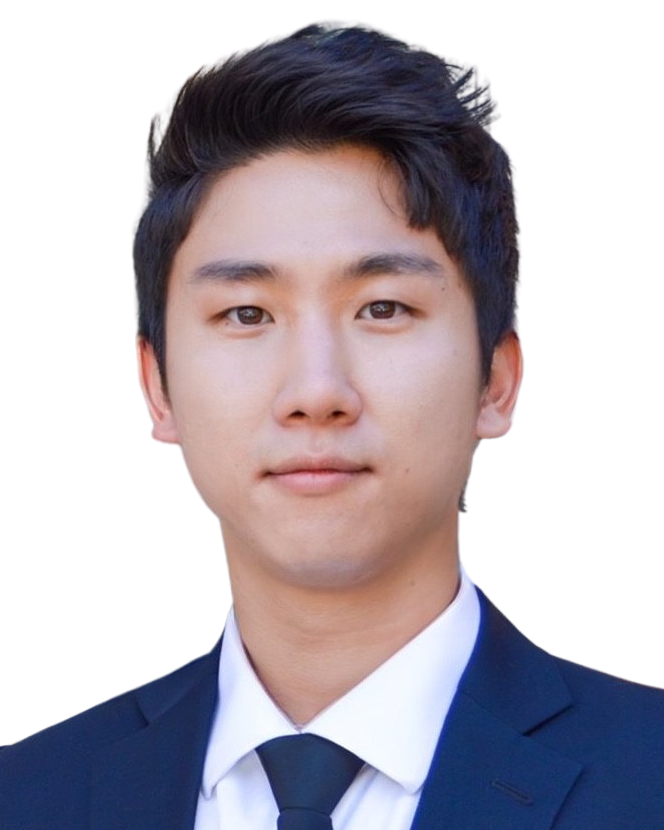}}]{Mincheol Kim}
has been working toward the Ph.D.
degree since 2018 at The
University of Texas at Austin, TX, USA.
His research interests include dexterous manipulation, reinforcement learning, optimal control, and mechanism design.
\end{IEEEbiography}


\begin{IEEEbiography}{Ashish D. Deshpande}
\end{IEEEbiography}





\end{document}